\pdfoutput=1

\documentclass[11pt]{article}

\usepackage{caption}       

\usepackage{array} 
\usepackage{enumitem}
\usepackage{threeparttable}  
\usepackage{tabularx}        
\usepackage{multirow}        

\usepackage[preprint]{acl}

\usepackage{times}
\usepackage{latexsym}
\usepackage[most]{tcolorbox}
\usepackage{listings}
\usepackage{booktabs} 
\usepackage{makecell}
\usepackage[T1]{fontenc}

\usepackage[utf8]{inputenc}

\usepackage{microtype}

\usepackage{inconsolata}
\usepackage{tcolorbox}
\usepackage{ragged2e}
\usepackage{graphicx}
\usepackage{natbib}
\usepackage{float}     
\usepackage{xcolor}

\tcbuselibrary{skins, breakable, theorems}

\title{Efficient Reasoning via Chain of Unconscious Thought}

\author{
  \textbf{Ruihan Gong}$^{1}$\thanks{Equal Contribution.} 
  \: \textbf{Yue Liu}$^{2}$\footnotemark[1] 
  \: \textbf{Wenjie Qu}$^{2}$ \\
  \: \textbf{Mingzhe Du}$^{3,2}$ 
  \: \textbf{Yufei He}$^{2}$
  \: \textbf{Yingwei Ma}$^{4}$
  \: \textbf{Yulin Chen}$^{2}$
  \: \textbf{Xiang Liu}$^{2}$\\
  \: \textbf{Yi Wen}$^{2}$
  \: \textbf{Xinfeng Li}$^{3}$
  \: \textbf{Ruidong Wang}$^{5}$
  \: \textbf{Xinzhong Zhu}$^{5}$ \\
  \: \textbf{Bryan Hooi}$^{2}$
  \: \textbf{Jiaheng Zhang}$^{2}$
  \\
  $^{1}$Huazhong University of Science and Technology \\
  $^{2}$National University of Singapore 
  $^{3}$Nanyang Technological University\\
  $^{4}$Moonshot AI
  $^{5}$Zhejiang Normal University\\
}

\begin{document}
\maketitle
\begin{abstract}
Large Reasoning Models (LRMs) achieve promising performance but compromise token efficiency due to verbose reasoning processes. 
Unconscious Thought Theory (UTT) posits that complex problems can be solved more efficiently through internalized cognitive processes.
Inspired by UTT, we propose a new reasoning paradigm, termed Chain of Unconscious Thought (CoUT), to improve the token efficiency of LRMs by guiding them to mimic human unconscious thought and internalize reasoning processes.
Concretely, we first prompt the model to internalize the reasoning by thinking in the hidden layer. 
Then, we design a bag of token-efficient strategies to further help models reduce unnecessary tokens yet preserve the performance. 
Our work reveals that models may possess beneficial unconscious thought, enabling improved efficiency without sacrificing performance.
Extensive experiments demonstrate the effectiveness of CoUT. Remarkably, it surpasses CoT by reducing token usage by 47.62\% while maintaining comparable accuracy, as shown in Figure \ref{fig:overall}. The code of CoUT is available at this link\footnote{https://github.com/Rohan-GRH/CoUT}. 




\end{abstract}

\begin{figure}[ht]
    \begin{minipage}{0.5\textwidth}
        \includegraphics[width=\linewidth]{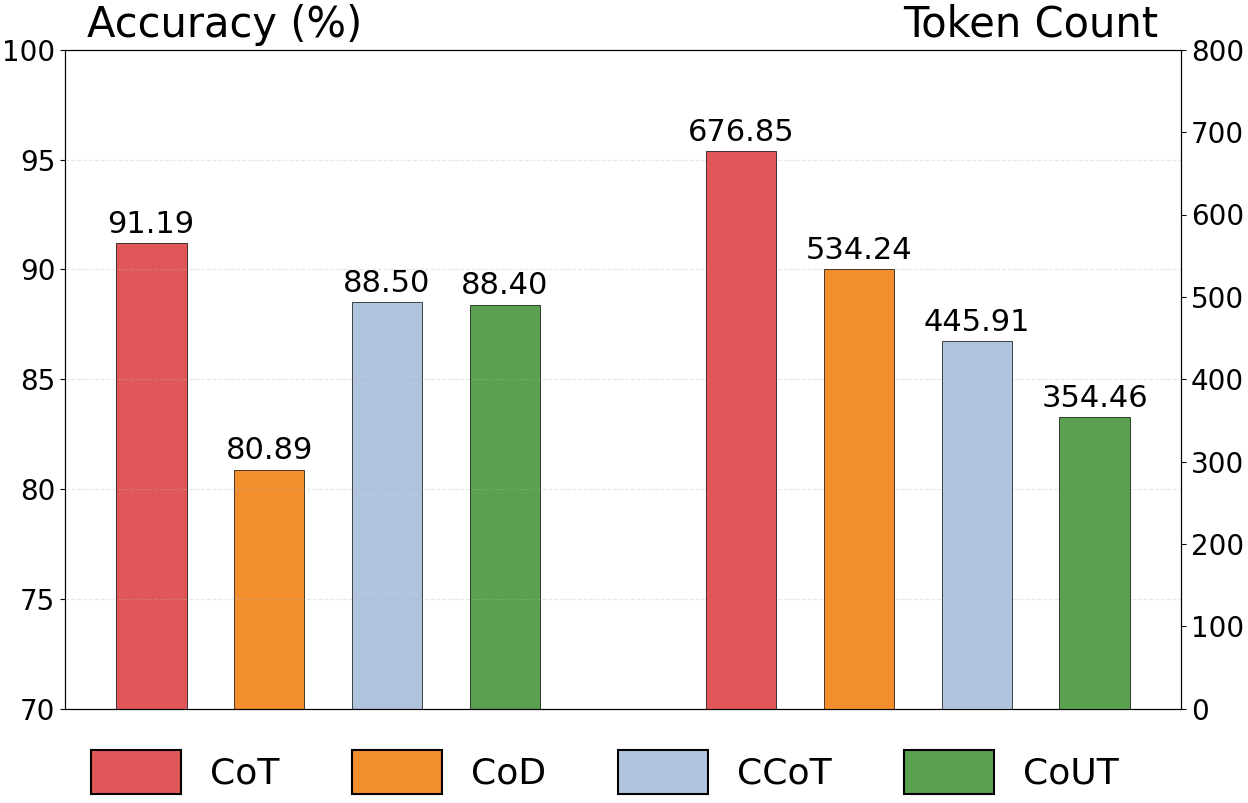}
        \captionsetup{width=0.95\linewidth} 
        \caption{\textbf{Average Performance and Tokens of CoUT and Baselines for 4 LRMs over 4 Benchmarks.}}
        \label{fig:overall}
    \end{minipage}
\end{figure}


\section{Introduction}

Large reasoning models (LRMs) \citep{o3o4mini,deepseek_r1} have demonstrated promising performance in complex tasks like code, math, and computer use via Chain of Thought (CoT) reasoning \citep{CoT}. 
Despite their effectiveness, LRMs are token-inefficient due to high token costs of the reasoning processes \citep{liu2025efficient}.


To alleviate this problem, the existing training-based methods are proposed via supervised fine-tuning \citep{kang2024c3ot}, reinforcement learning \citep{luo2025o1}, latent CoT \citep{hao2024training}, etc. Although effective, they require further training on LRMs, leading to high inference costs. 
Differently, the more adaptable training-free methods are proposed with prompt engineering strategies \citep{nayab2024concise, xu2025chain}, reasoning delegation approaches \citep{aytes2025sketch}, and dynamic optimization methods \citep{sui2025meta}. 
Despite these advancements, existing methods still generate explicit and redundant reasoning processes, leading to token inefficiency, as shown in Figure \ref{fig:overall}. 






To solve this problem, we introduce Unconscious Thought Theory (UTT) from cognitive science, which suggests complex problems can be solved more efficiently through internalized cognitive processes. 
From this principle, we propose Chain of Unconscious Thought (CoUT), a novel paradigm that encourages models to conduct the reasoning process within their hidden layers.
Concretely, it first prompts the model to internalize reasoning processes without emitting detailed chains, thereby achieving significant reasoning compression.
In addition, we introduce a bag of token-efficient strategies to minimize the unnecessary token costs while preserving reasoning accuracy. 
In this manner, CoUT significantly reduces the explicit token outputs required during inference while maintaining or improving accuracy. 
To evaluate the effectiveness of CoUT, we conduct extensive experiments on a wide range of mathematical reasoning benchmarks, including both open-ended and multiple-choice questions. These results underscore the potential of leveraging unconscious thought paradigms to enhance efficiency. 
Extensive experiments demonstrate the effectiveness of CoUT. 
As shown in Figure \ref{fig:overall}, it notably reduces token usage by 20.51\% with only a 0.1\% drop in accuracy, outperforming the runner-up on average. 
The main contributions are summarized as follows.

\begin{itemize}[leftmargin=1.5em, itemsep=0.4em, parsep=0em]
\item We introduce UTT and propose CoUT, a new reasoning paradigm, to improve the token efficiency of LRMs by internalizing the reasoning.
\item We design a bag of token-efficient strategies to help models reduce unnecessary tokens while preserving reasoning performance.
\item Extensive experiments and analyses demonstrate the effectiveness and efficiency of CoUT.
\end{itemize}


\section{Related Work}

\subsection{Reasoning Ability LRMs}

Reasoning capabilities are vital for LRMs, with significant research devoted to enhancing these abilities. Pioneering work \citep{CoT,kojima2022large} introduced step-by-step thinking through prompting. Additional frameworks like self-correction \citep{self_correct_1}, self-critique \citep{self_critique}, debate \citep{debate_1,debate_2}, and plan-and-solve \citep{wang2023plan} have further advanced reasoning capacities. \citet{ma2023training} investigates code data's impact on LRMs reasoning during training.

OpenAI's o1 model demonstrated enhanced reasoning through test-time scaling, inspiring similar models like QwQ \citep{qwq}, QvQ \citep{qvq}, DeepSeek \citep{deepseek_r1}, and Kimi \citep{kimi2025kimi}. Furthermore, OpenAI's o3 and o4-mini \citep{o3o4mini} has shown promising results on the ARG-AGI benchmark \citep{ARC_AGI}. 
LLMs progressively shift from intuitive processing (System 1) to deliberative reasoning (System 2) \citep{li2025system}. Besides, researchers demonstrate that reasoning can improve safety \citep{liuyue_GuardReasoner,liuyue_GuardReasoner-VL} and alleviate hallucination \citep{gao2025exploring}. However, \cite{overthinking} examines the overthinking problem observed in o1-like models. To alleviate this problem, token efficiency methods \citep{liu2025efficient} are proposed to reduce the token costs while maintaining the reasoning quality. 


\subsection{Token Efficiency of LRMs}

Token efficiency remains a key challenge for LRMs \citep{overthinking}, as reasoning methods boost performance but increase inference costs \citep{liu2025efficient}. Recent token-efficient approaches can be categorized into two classes, including training-based methods and training-free methods.

Training-based methods require substantial computational resources for fine-tuning or reinforcement learning. These include supervised approaches like C3oT \citep{kang2024c3ot}, which fine-tunes on condensed reasoning chains, and TokenSkip \citep{xia2025tokenskip}, which prunes token-by-token based on importance. Reinforcement learning approaches like Kimi k1.5 \citep{kimi2025kimi} and O1-Pruner \citep{luo2025o1} integrate length-based rewards to discourage verbosity. Implicit latent CoT methods like COCONUT \citep{hao2024training} and CCoT \citep{cheng2024compressed} encode reasoning in hidden representations rather than explicit tokens.

In contrast, training-free methods can be applied directly at inference time without additional training costs. CCoT \citep{nayab2024concise} and CoD \citep{xu2025chain} use prompt engineering to confine reasoning to essential steps. SoT \citep{aytes2025sketch} employs a smaller router model to generate concise reasoning sketches, while Meta-Reasoner \citep{sui2025meta} applies a contextual multi-armed bandit to dynamically optimize efficiency. 
Our proposed CoUT is a training-free reasoning paradigm. Unlike the existing methods, CoUT improves the token efficiency of LRMs by guiding them to mimic human unconscious thought and internalize reasoning processes.



\section{Chain of Unconscious Thought}

This section introduces our proposed Chain of Unconscious Thought (CoUT). First, we give the problem definition and analyze the limitations of the existing training-free reasoning paradigms. Then, we introduce the Unconscious Thought Theory (UTT). Then, based on UTT, we present two components in CoUT, including Reasoning Process Internalization (RPI) and Token-Efficient Strategies (TES).

\subsection{Problem Definition}
Given a user's query $\mathcal{Q}$, the LRMs $\mathcal{M}$ will output the reasoning process $\mathcal{R}$ and the predicted final answer $\hat{\mathcal{Y}}$, i.e., $\{\mathcal{R}, \hat{\mathcal{Y}} \} = \mathcal{M}(\mathcal{Q})$. The predicted final answer will be compared with the ground truth $\mathcal{Y}$ to evaluate the performance, i.e., $s = \text{eval}(\hat{\mathcal{Y}},\mathcal{Y})$, where $\text{eval}$ denotes the evaluation method and $s$ denote the model performance. This paper aims to optimize the reasoning process by minimizing its length $\text{len}(\mathcal{R})$ and simultaneously maximizing the performance score $s$ by designing novel prompting strategies, i.e., $\min_Q \text{len}(\mathcal{R}),\ \max_Q s$.

\subsection{Limitations of Existing Methods}

The token efficiency of the recent training-free reasoning paradigms is limited. We introduce them and analyze their underlying limitations as follows.


\begin{tcolorbox}[
    colback=gray!10, 
    colframe=gray!70,
    title={\Centering Chain-of-Thought}, 
    boxrule=0.5pt,
    arc=2pt,
    fonttitle=\bfseries,
    coltitle=white!50,
    listing only,
    listing options={
        basicstyle=\ttfamily\small,
        breaklines=true,
        escapeinside=@@,
    }
]
Think step by step to answer the following question.
\end{tcolorbox}

\textbf{Chain-of-Thought (CoT)} \citep{CoT} improves reasoning accuracy by forcing models to ``think step by step'', but may generate unnecessarily verbose outputs that consume substantial token budgets.
This inefficiency stems from fully externalizing every reasoning step regardless of importance. 



\begin{tcolorbox}[
    colback=gray!10, 
    colframe=gray!70,
    title={\Centering Chain-of-Draft}, 
    boxrule=0.5pt,
    arc=2pt,
    fonttitle=\bfseries,
    coltitle=white!50,
    listing only,
    listing options={
        basicstyle=\ttfamily\small,
        breaklines=true,
        escapeinside=@@,
    }
]
Think step by step, but only keep minimum draft for each thinking step, with 5 words at most.
\end{tcolorbox}

\textbf{Chain-of-Draft (CoD)} \citep{xu2025chain} constrains each reasoning step to five words. However, this prompting strategy has limited adaptability, as the complexity of reasoning steps inherently depends on the task. Moreover, it may not effectively reduce token costs, as the number of steps could increase to compensate for brevity in each step.

\begin{tcolorbox}[
    colback=gray!10, 
    colframe=gray!70,
    title={\Centering Concise Chain-of-Thought}, 
    boxrule=0.5pt,
    arc=2pt,
    fonttitle=\bfseries,
    coltitle=white!50,
    listing only,
    listing options={
        basicstyle=\ttfamily\small,
        breaklines=true,
        escapeinside=@@,
    }
]
Let's think step by step and limit the answer length to 45 words.
\end{tcolorbox}

\textbf{Concise Chain-of-Thought (CCoT)} \citep{nayab2024concise} constrains reasoning to 45 tokens while leaving answers unconstrained. Its limitations are similar to CoD, such as limited adaptability to varying task complexity. Moreover, it may fail to significantly reduce overall token usage when tasks require longer answers or additional reasoning steps to compensate for the strict constraint.


\begin{tcolorbox}[
    colback=gray!10, 
    colframe=gray!70,
    title={\Centering Token-Budget-Aware LLM Reasoning}, 
    boxrule=0.5pt,
    arc=2pt,
    fonttitle=\bfseries,
    coltitle=white!50,
    listing only,
    listing options={
        basicstyle=\ttfamily\small,
        breaklines=true,
        escapeinside=@@,
    }
]
Let's think step by step and use less than {budget} tokens.
\end{tcolorbox}

\textbf{Token-Budget-Aware Prompt (TALE-EP)} \citep{2412.18547} predicts token budgets before generating answers, reducing costs through planned allocation.
However, this two-stage method may cost more tokens as, the model's initial token prediction can sometimes be several times more than what is needed for the final answer.

The common limitation across these methods is their reliance on externalized reasoning—converting complex cognitive processes into sequences of tokens. Whether through full verbalization (CoT), word-limited steps (CoD), token-constrained reasoning (CCoT), or budget-aware generation (TALE-EP), all these approaches mandate that reasoning steps appear in the output. 



\subsection{Unconscious Thought Theory}

UTT \citep{UTT} distinguishes between two modes of thinking: conscious thought and unconscious thought. Unconscious thought operates without the constraints of working memory capacity, allowing it to process larger volumes of information and making it particularly suitable for handling complex decision-making tasks. Differently, conscious thought performs better when addressing simple, rule-based problems where focused attention is beneficial.

One approach \citep{both} to improve the adaptability is to combine unconscious thought and conscious thought. Building upon this theoretical foundation, we propose CoUT. Our approach consists of two complementary components. The first component, Reasoning Process Internalization (RPI), encourages models to internalize their reasoning within their hidden layers rather than explicitly generating each step of the reasoning process as output tokens. The second component, Token-Efficient Strategies (TES), addresses the inevitable output of some reasoning processes by the model. Through TES, we implement effective strategies that reduce the number of tokens generated by the model without compromising accuracy.

\subsection{Reasoning Process Internalization}
Inspired from UTT, which posits that complex problems can be solved efficiently through internalized cognitive processes, we propose a fundamentally different approach. Rather than externalizing every reasoning step, we guide the model to perform reasoning implicitly within its hidden layers.
It is implemented via two key prompting strategies:

\begin{itemize}[leftmargin=1.5em, itemsep=0.3em, parsep=0em]

\item \textbf{Hidden Layer Processing}: We explicitly instruct the model: \textit{``Process and solve problems fully in your hidden layer thinking.''} This directive encourages the model to utilize its internal computational capacity for reasoning without converting intermediate steps into tokens.
\item \textbf{Minimal Output Constraint}: We further guide the model with: \textit{"Output bare minimum answers with only single-line reasoning when necessary for clarity."} This establishes an expectation of conciseness while allowing minimal articulation when essential for accuracy.
\end{itemize}

These instructions leverage the model's ability to perform complex reasoning within its parameter space, the "unconscious" processing capacity that exists prior to token generation. By encouraging the model to leverage this capability, we reduce the cognitive-linguistic conversion overhead that characterizes traditional reasoning approaches.


\begin{tcolorbox}[colback=blue!5!white, colframe=blue!60!black,breakable, title=\Centering Reasoning Process Internalization]
    Process and solve problems fully in your hidden layer thinking. Output bare minimum answers with only single-line reasoning when necessary for clarity.
\end{tcolorbox}

\subsection{Token-Efficient Strategies}
To further maximize token efficiency without sacrificing accuracy, we implement a comprehensive set of token conservation strategies through carefully crafted prompt engineering:

\begin{itemize}[leftmargin=1.5em, itemsep=0.3em, parsep=0em]
\item \textbf{Token Conservation Framing}: We establish token efficiency as a priority by beginning instructions with \textit{"TOKEN CONSERVATION MODE ACTIVE"} and \textit{"You are running on a system with severe token limitations."}
\item \textbf{Symbol Usage}: We encourage efficient representation through \textit{"Use symbols/abbreviations when clear (e.g., \&, w/, =, →),"} using symbolic notation's natural efficiency advantage.
\item \textbf{Language Streamlining}: We direct the model to \textit{"Omit articles when meaning remains clear"} and \textit{"Strip all non-essential words,"} eliminating common sources of token waste.
\item \textbf{Efficiency-Accuracy Balance}: We quantify priorities with \textit{"Each saved token equals +1 efficiency point while each accuracy error costs -100 efficiency points,"} establishing that accuracy remains paramount in token constraints.
\item \textbf{Minimal Precision}: We reinforce the objective with \textit{"Focus exclusively on maximum precision with minimum verbosity,"} ensuring token reduction does not compromise performance.
\end{itemize}

\begin{tcolorbox}[colback=blue!5!white, colframe=blue!60!black,breakable, title=\Centering Token-Efficient Strategies]
    1. TOKEN CONSERVATION MODE ACTIVE. \\ 
    2. You are running on a system with severe token limitations. \\
    3. Use symbols/abbreviations when clear (e.g., \&, w/, =, →). \\
    4. Omit articles (a, an, the) when meaning remains clear. \\
    5. Strip all non-essential words including greetings, acknowledgments, and explanations. \\
    6. Each saved token equals +1 efficiency point while each accuracy error costs -100 efficiency points. \\
7. Focus exclusively on maximum precision with minimum verbosity.
\end{tcolorbox}

By combining these two components—internal reasoning processes and efficient output strategies—CoUT achieves significant reductions in token usage while maintaining - and, in some cases, improving -reasoning accuracy.

\begin{tcolorbox}[colback=blue!5!white, colframe=blue!60!black,breakable, title=\Centering Chain of Unconscious Thought]
  1. You are running on a system with severe token limitations.\\
  2. Process and solve problems fully in your hidden layer thinking.\\
  3. Output bare minimum answers with only single-line reasoning when necessary for clarity.\\
  4. Use symbols/abbreviations when clear (e.g., \&, w/, =, →).\\
  5. Omit articles (a, an, the) when meaning remains clear.\\
  6. Strip all non-essential words including greetings, acknowledgments, and explanations.\\
  7. Each saved token equals +1 efficiency point while each accuracy error costs -100 efficiency points.\\
  8. Focus exclusively on maximum precision with minimum verbosity.
\end{tcolorbox}

\begin{table*}[!t]
\renewcommand{\arraystretch}{1.}
\centering
\caption{\textbf{Performance (\%) and Token Costs of Our Proposed CoUT and Baselines for 4 LRMs on 4 Benchmarks.} The \textbf{bold} values and \underline{underlined} values denote the best and the runner-up token efficiency, respectively. }
\label{tab:all_results}
\setlength{\tabcolsep}{3pt}
\resizebox{1.0\linewidth}{!}{
\begin{tabular}{ccccccccccc}
\toprule
\multirow{2}{*}{\textbf{Method}} 
  & \multicolumn{2}{c}{\textbf{GPT-4o}} 
  & \multicolumn{2}{c}{\textbf{Claude 3.5 Sonnet}} 
  & \multicolumn{2}{c}{\textbf{O3-mini}} 
  & \multicolumn{2}{c}{\textbf{QWQ-32B}} 
  & \multicolumn{2}{c}{\textbf{Average}} \\
 & \textbf{Accuracy} & \textbf{Token} 
 & \textbf{Accuracy} & \textbf{Token} 
 & \textbf{Accuracy} & \textbf{Token} 
 & \textbf{Accuracy} & \textbf{Token} 
 & \textbf{Accuracy} & \textbf{Token} \\
\midrule
\multicolumn{11}{c}{GSM8K} \\
\midrule
CoT & 96.00 & 274.30 & 91.00 & 244.59 & 96.00 & 421.73 & 94.00 & 1561.91 & 94.25 & 625.63 \\
CoD & 85.90 & 77.80 & 67.70 & 74.72 & 95.37 & 805.90 & 52.00 & 682.41 & 75.24 & 410.21 \\
CCoT & 94.10 & 91.25 & 89.39 & 137.66 & 95.00 & 669.42 & 88.00 & 451.49 & 91.62 & \underline{337.46} \\
TALE-EP & 93.10 & 98.90 & 94.69 & 139.32 & 94.00 & 2414.43 & 78.00 & 4324.68 & 89.95 & 1744.33 \\

CoUT & 93.40 & 56.80 & 95.00 & 69.10 & 91.40 & 245.98 & 96.00 & 472.81 & 93.95 & \textbf{211.17} \\

\midrule
\multicolumn{11}{c}{SVAMP} \\
\midrule
CoT & 95.00 & 195.15 & 89.00 & 210.59 & 91.00 & 485.05 & 96.00 & 404.09 & 92.75 & 323.72 \\
CoD & 93.90 & 43.30 & 92.20 & 63.39 & 93.00 & 893.01 & 56.00 & 650.83 & 83.78 & 412.63 \\
CCoT & 93.70 & 72.13 & 91.80 & 100.45 & 94.00 & 583.39 & 87.00 & 388.24 & 91.63 & \underline{286.05} \\
TALE-EP & 94.40 & 63.38 & 93.90 & 82.33 & 100.00 & 2485.59 & 98.00 & 3832.05 & 96.58 & 1615.84 \\

CoUT & 90.07 & 48.37 & 93.90 & 42.59 & 94.00 & 185.26 & 97.00 & 400.91 & 94.60 & \textbf{169.21} \\
\midrule

\multicolumn{11}{c}{MathQA} \\
\midrule
CoT & 86.00 & 417.86 & 90.00 & 295.55 & 90.00 & 679.76 & 87.00 & 2342.8 & 88.25 & 933.99 \\
CoD & 77.40 & 156.90 & 68.84 & 81.41 & 90.00 & 804.23 & 93.00 & 1530.42 & 82.31 & 643.24 \\
CCoT & 83.00 & 187.20 & 80.44 & 136.87 & 94.00 & 588.58 & 90.00 & 1281.72 & 86.86 & \underline{548.59} \\
TALE-EP & 81.98 & 149.18 & 83.65 & 152.03 & 94.00 & 2294.09 & 94.00 & 5433.69 & 88.41 & 2007.25 \\

CoUT & 81.20 & 149.40 & 69.92 & 73.03 & 90.00 & 408.36 & 92.00 & 1439.58 & 83.28 & \textbf{517.59} \\
\midrule
\multicolumn{11}{c}{AQuA} \\
\midrule
CoT & 88.00 & 425.03 & 86.00 & 291.61 & 95.0 & 652.72 & 89.00 & 1926.78 & 89.50 & 824.04 \\
CoD & 73.60 & 255.60 & 69.30 & 83.89 & 97.00 & 874.45 & 89.00 & 1469.64 & 82.23 & 670.90 \\
CCoT & 81.90 & 164.85 & 80.70 & 132.19 & 88.00 & 708.75 & 85.00 & 1440.44 & 83.90 & \underline{611.56} \\
TALE-EP & 83.46 & 202.49 & 79.53 & 186.75 & 89.00 & 2164.37 & 87.00 & 5259.03 & 84.75 & 1953.16 \\

CoUT & 80.30 & 137.60 & 74.80 & 74.52 & 80.00 & 449.41 & 92.00 & 1417.91 & 81.78 & \textbf{519.86} \\

\bottomrule
\end{tabular}}
\end{table*}

\section{Experiments}

This section evaluates the effectiveness of CoUT in reducing tokens while maintaining performance.

\subsection{Experimental Setup}

We compare CoUT with baseline methods, including CoT \citep{CoT}, CoD \citep{xu2025chain}, CCoT \citep{2401.05618}, and TALE-EP\citep{2412.18547}, on several reasoning tasks. For TALE-EP, the recorded token count represents the sum of tokens from the model’s two-round responses (first for estimating required tokens, then for generating the constrained answer). In this study, all experiments are conducted under a zero-shot learning setup, meaning that the models do not receive any training or fine-tuning on the specific datasets used for evaluation.

\textbf{CoT }is a prompting methodology that encourages language models to decompose complex reasoning tasks into a series of intermediate steps. By instructing the model to "think step by step," 

\textbf{CoD }maintains the step-by-step reasoning of CoT but constrains each step to a maximum of five words. This brevity reduces token usage and response time while preserving reasoning accuracy.

\textbf{CCoT } is a variant of CoT that limits the reasoning process to 45 tokens, promoting concise responses. The final answer remains unconstrained, balancing brevity with completeness.

\textbf{TALE-EP }is a two-step strategy designed to optimize token usage. First, the model estimates the number of tokens required to answer a question. Then, it generates a response within this predicted budget, effectively reducing token costs while maintaining accuracy.

\textbf{CoUT} boosts reasoning efficiency by internalizing logic in hidden layers, avoiding explicit steps. It employs token-efficient strategies to cut costs while retaining reasoning quality.


Our experiments evaluate the performance of four leading large language models on four math reasoning datasets: GPT-4o (gpt-4o-2024-0806) from OpenAI, Claude 3.5 Sonnet (claude3-5-sonnet-20240620) from Anthropic, O3-mini, and Qwen (QwQ-32B), representing a mix of top-tier proprietary models and strong open-source alternatives. We evaluate CoUT on 4 math datasets:

\begin{itemize}
\setlength{\itemsep}{-0.4em}
    \item \textbf{GSM8K \citep{2110.14168}}: A dataset of grade-school-level word problems covering arithmetic, algebra, and logic.
    \item \textbf{SVAMP \citep{patel-etal-2021-nlp}}: A dataset of multi-step word problems that require reasoning over multiple pieces of information.
    \item \textbf{MathQA \citep{jie-etal-2024-reft}}: This dataset tests the models' ability to solve math problems combining arithmetic and algebra.
    \item \textbf{Aqua \citep{huang_automated_2022}}: A dataset containing multi-step arithmetic reasoning tasks designed to evaluate reasoning abilities.
\end{itemize}

\subsection{Comparison Experiments}

This section compares the performance and token costs of our proposed method with the baselines. It mainly contains two aspects, including arithmetic reasoning and mathematical reasoning.

\subsubsection{Arithmetic Reasoning}

We analyze the performance of large language models on arithmetic reasoning tasks using GSM8K and SVAMP datasets, which evaluate models' capabilities in handling multi-step mathematical problems.

As shown in Table~\ref{tab:all_results}, when averaging results across both datasets: 
($\mathrm{I}$) The CoUT method demonstrates significantly lower token consumption (190.19) compared to other methods, being approximately 88.7\% less than TALE-EP (1680.09) and 39.0\% less than CCoT (311.76).
($\mathrm{II}$) Despite this substantial reduction in token usage, CoUT maintains high accuracy (94.28\%), which is comparable to CoT (93.50\%) and outperforms both CoD (79.51\%) and CCoT (91.63\%).

Experimental results indicate that for arithmetic reasoning tasks, the CoUT method effectively reduces token output while maintaining high accuracy, significantly improving reasoning efficiency and cost-effectiveness in practical applications.

\subsubsection{Mathematical Reasoning}

MathQA and AQUA datasets are categorized as mathematical reasoning due to their requirement for sophisticated symbolic manipulation, algebraic transformations, and complex multi-step logical inference processes.

Table~\ref{tab:all_results} reveals compelling efficiency advantages for the CoUT method in these complex reasoning tasks. ($\mathrm{I}$) CoUT consumes just 518.73 tokens on average—more than 100 tokens fewer than the next most efficient approach (CCoT at 580.08). ($\mathrm{II}$) On the accuracy front, CoUT (82.53\%) outperforms CoD (82.27\%) while maintaining reasonable performance relative to other methods, despite its significantly reduced computational footprint.

These results demonstrate that even in complex reasoning tasks, CoUT can effectively reduce large model output tokens without significantly compromising accuracy.

\subsection{Ablation Studies}

\begin{figure*}[!t]
    \begin{minipage}{1.0\textwidth}
        \includegraphics[width=\linewidth]{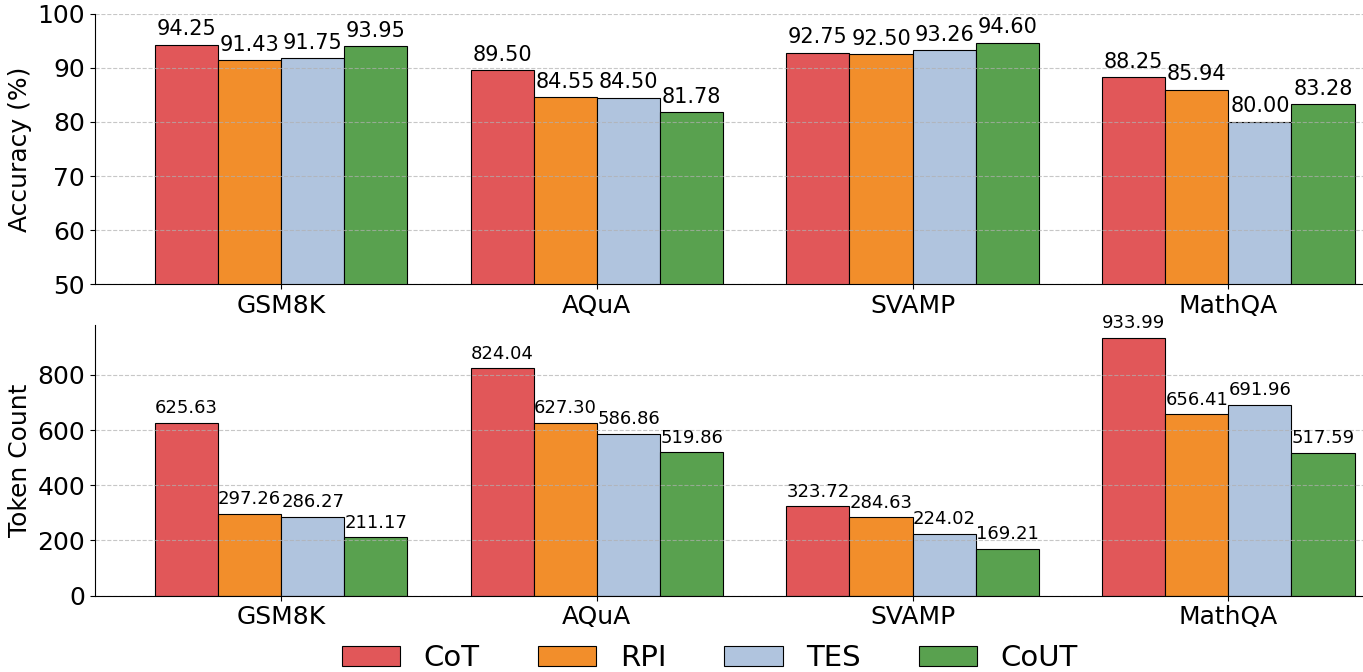}
        \captionsetup{width=0.95\linewidth} 
        \caption{\textbf{Ablation Studies on Our Proposed CoUT.} CoT denotes the Chain-of-Thought baseline. RPI denotes Reasoning Process Internalization. TES denotes Token-Efficient Strategies. }
        \label{fig:ablation}
    \end{minipage}
\end{figure*}

\begin{table}[ht]
\renewcommand{\arraystretch}{1.}
\centering
\caption{\textbf{Statistics of Benchmarks.}}
\label{tab:bench}
\setlength{\tabcolsep}{3pt}
\resizebox{1.0\linewidth}{!}{
\begin{tabular}{cccc}
\toprule
& \textbf{Benchmark} & \textbf{Num. of Sample} & \textbf{Task Type} \\
\midrule
\multicolumn{4}{c}{Arithmetic Reasoning} \\ \midrule
& GSM8K              & 1319              & Open-ended QA                    \\
& SVAMP              & 1000              & Open-ended QA                    \\
\addlinespace
\midrule
\multicolumn{4}{c}{Mathematical Reasoning} \\ \midrule
& AQUA               & 254               & Multi-choice QA                    \\
& MATHQA             & 2985              & Multi-choice QA                    \\
\bottomrule
\end{tabular}
}
\end{table}

This section verifies the effectiveness of components in our Chain of Underspecified Thought (CoUT) method. As shown in Figure~\ref{fig:ablation}, we conduct ablation studies on four models (GPT-4o, Claude 3.5 Sonnet, O3-mini, and QWQ-32B) across four mathematical reasoning datasets (GSM8K, AQuA, SVAMP, and MathQA).

First, we evaluate the individual components of CoUT. "RPI" denotes Reasoning Process Internalization, which encourages reasoning to occur within the model's hidden layers. "TES" denotes our Token-Efficiency Strategies, which implements techniques to reduce output verbosity. "CoT" represents traditional Chain of Thought prompting with explicit reasoning steps. "CoUT" combines both RPI and TES approaches. 

We have the following conclusions: ($\mathrm{I}$) The RPI component achieves 88.61\% accuracy with 466.40 tokens, reducing token consumption by 31.1\% compared to CoT while maintaining comparable performance. ($\mathrm{II}$) The TES component reaches 87.38\% accuracy with 447.28 tokens, demonstrating a 33.9\% reduction in token usage. ($\mathrm{III}$) Each component individually contributes to performance improvement. And the combination achieves the best performance. 

Second, by combining both components, CoUT achieves 88.40\% accuracy with only 354.46 tokens consumed, representing a 47.6\% reduction in token usage compared to CoT's 676.85 tokens, while the accuracy decrease is limited to 2.79\% (from 91.19\% to 88.40\%). This demonstrates that CoUT achieves the best token efficiency, showing the effectiveness of integrating both components.

\subsection{Case Studies}

To illustrate the contrast between CoT \citep{CoT}, and CoD \citep{xu2025chain}  prompting with CoUT prompting, we selected a representative example from our experiments. 

\begin{tcolorbox}[
    colback=gray!10, 
    colframe=gray!70,
    title={\Centering Question}, 
    boxrule=0.5pt,
    arc=2pt,
    fonttitle=\bfseries,
    coltitle=white!50,
    listing only,
     breakable,
    listing options={
        basicstyle=\ttfamily\small,
        breaklines=true,
        escapeinside=@@,
    }
]
\ttfamily
{Q:} The difference between simple interest and C.I. at the same rate for Rs.5000 for 2 years in Rs.72. The rate of interest is?

Options:\\
A. 10\%  \\
B. 12\%  \\
C. 6\%  \\
D. 8\%  \\
E. 4\%
\end{tcolorbox}

As seen in the following example, the CoT response is highly verbose, meticulously documenting each step of the reasoning process. While this approach provides complete transparency of the model's thinking, it consumes substantial computational resources unnecessarily. The exhaustive calculations and explanations result in significant token usage even for problems that don't require such detailed exposition.

\begin{tcolorbox}[
    colback=gray!10, 
    colframe=gray!70,
    title={\Centering Chain-of-Thought}, 
    boxrule=0.5pt,
    arc=2pt,
    fonttitle=\bfseries,
    coltitle=white!50,
    listing only,
     breakable,
    listing options={
        basicstyle=\ttfamily\small,
        breaklines=true,
        escapeinside=@@,
    }
]
Answer: \\
To solve this problem, we need to find the rate of interest where the difference between the compound interest (C.I.) and simple interest (S.I.) for 2 years is Rs. 72 on a principal amount of Rs. 5000.

Let's denote the rate of interest as R\%.

1. \textbf{Simple Interest (S.I.) for 2 years:}
\[
\text{S.I.} = \frac{P \times R \times T}{100} = \frac{5000 \times R \times 2}{100} = 100R
\]

2. \textbf{Compound Interest (C.I.) for 2 years:}
The formula for compound interest is:
\[
\text{C.I.} = P \left(1 + \frac{R}{100}\right)^T - P
\]
For 2 years, it becomes:
\[
\text{C.I.} = 5000 \left(1 + \frac{R}{100}\right)^2 - 5000
\]

3. \textbf{Difference between C.I. and S.I.:}
Given that the difference is Rs. 72:
\[
\text{C.I.} - \text{S.I.} = 72
\]
Substituting the expressions for C.I. and S.I.:
\[
5000 \left(1 + \frac{R}{100}\right)^2 - 5000 - 100R = 72
\]
Simplifying:
\[
5000 \left(1 + \frac{R}{100}\right)^2 - 5000 = 100R + 72
\]
\[
5000 \left(1 + \frac{R}{100}\right)^2 = 100R + 5072
\]
\[
\left(1 + \frac{R}{100}\right)^2 = \frac{100R + 5072}{5000}
\]

4. \textbf{Solving for R:}
Let's try the options to find the correct R.

- \textbf{Option A: 10\%}
\[
\left(1 + \frac{10}{100}\right)^2 = \left(1.1\right)^2 = 1.21
\]
\[
5000 \times 1.21 = 6050
\]
\[
\text{C.I.} = 6050 - 5000 = 1050
\]
\[
\text{S.I.} = 100 \times 10 = 1000
\]
\[
\text{Difference} = 1050 - 1000 = 50 \neq 72
\]

- \textbf{Option B: 12\%}
\[
\left(1 + \frac{12}{100}\right)^2 = \left(1.12\right)^2 = 1.2544
\]
\[
5000 \times 1.2544 = 6272
\]
\[
\text{C.I.} = 6272 - 5000 = 1272
\]
\[
\text{S.I.} = 100 \times 12 = 1200
\]
\[
\text{Difference} = 1272 - 1200 = 72
\]

The correct rate of interest is 12\%.

\end{tcolorbox}

The CoD method shows improvement in conciseness compared to CoT, condensing the solution into key steps. However, it still includes multiple reasoning steps that might be redundant for relatively straightforward problems. The structured approach, while more efficient than CoT, continues to externalize computational steps that could potentially be handled internally by the model.

In contrast, the CoUT response demonstrates remarkable efficiency by directly providing the correct answer without unnecessary elaboration. For this problem, CoUT correctly identifies that "12\%" is the solution while using dramatically fewer tokens. It showcases CoUT's ability to leverage the model's internal reasoning capabilities without forcing every step to be externalized in text.

\begin{tcolorbox}[
    colback=gray!10, 
    colframe=gray!70,
    title={\Centering Chain-of-Draft}, 
    boxrule=0.5pt,
    arc=2pt,
    fonttitle=\bfseries,
    coltitle=white!50,
    listing only,
    listing options={
        basicstyle=\ttfamily\small,
        breaklines=true,
        escapeinside=@@,
    }
]
Answer: \\
1. Formula: CI - SI = Rs.72  \\
2. Principal = Rs.5000, Time = 2 years  \\
3. CI formula: \( P(1 + r/100)^n \) \\
4. SI formula: \( P \times r \times n / 100 \) \\
5. Calculate: \( (1 + r/100)^2 - 1 = 72/5000 \) \\
6. Simplify: \( r^2 / 10000 = 72 / 5000 \) \\
7. Solve: \( r^2 = 144 \) \\
8. Result: \( r = 12 \) 
\end{tcolorbox}

Our method fundamentally changes how reasoning is performed with language models. Rather than forcing every computational step to be externalized in text, CoUT allows the model to leverage its internal neural mechanisms for straightforward calculations and reasoning steps. The model only outputs information when it's necessary to clarify its thought process or when explicitly requested.


\begin{tcolorbox}[colback=blue!5!white, colframe=blue!60!black,breakable, title=\Centering Chain of Unconscious Thought]
Answer: \\
Rate = 12\%
\end{tcolorbox}

\section{Conclusion}

Inspired by Unconscious Thought Theory (UTT), we introduce Chain of Unconscious Thought (CoUT), a novel approach consisting of two complementary components. The first component, Reasoning Process Internalization (RPI), stimulates the implicit reasoning capabilities of large language models, encouraging them to minimize their output by performing simpler reasoning steps internally. The second component, Token-Efficient Strategies (TES), effectively compresses model outputs without compromising answer accuracy, significantly reducing token generation while maintaining response quality. This combined approach enables models to respond to queries more efficiently. Extensive experiments and analyses demonstrate the effectiveness and efficiency of our proposed CoUT. Future work will extend CoUT to diverse reasoning domains, including commonsense tasks and code tasks, explore few-shot learning scenarios, and evaluate its effectiveness on larger models to quantify potential efficiency gains at scale. Besides, the few token costs may bring safety concerns for LRMs \citep{wang2025safety,wang2025comprehensive}. It is worthy developing safe yet efficient methods for LRMs. 

\section{Limitaitons}

Although CoUT show significant improvements in token efficiency and accuracy across various models, there are some limitations when applied to specific large models like Qwen/QwQ-32B. During experiments with Qwen/QwQ-32B, we observed that no matter what prompt was used, the model’s response token count remained consistently high, making it difficult to significantly reduce the number of tokens required for generating a response.

Despite this challenge, CoUT still demonstrates the most balanced performance across the four datasets in Table~\ref{tab:all_results}. For example, on GSM8K, CoUT achieves an impressive 96\% accuracy while using only 472.81 tokens—significantly outperforming CoD which uses 45\% more tokens yet achieves only 52\% accuracy. This pattern repeats across datasets, with CoUT consistently delivering the best accuracy-to-token ratio.
What's particularly interesting is comparing these results with the CoT baseline. On SVAMP, for instance, CoUT achieves 97\% accuracy with 400.91 tokens, which is almost identical to CoT's token count (404.09) but with slightly better accuracy. This suggests that on certain tasks, CoUT can match or even exceed the performance of explicit reasoning while maintaining similar token efficiency.



\bibliography{0_arxiv}

\appendix

\section{Case Studies}

Case 1 from GSM8K dataset is demonstrated as follows. 
\begin{tcolorbox}[
    colback=gray!10, 
    colframe=gray!70,
    title={\Centering Question}, 
    boxrule=0.5pt,
    arc=2pt,
    fonttitle=\bfseries,
    coltitle=white!50,
    listing only,
     breakable,
    listing options={
        basicstyle=\ttfamily\small,
        breaklines=true,
        escapeinside=@@,
    }
]
\ttfamily
{Q:} Each pack of dvds costs 76 dollars. If there is a discount of 25 dollars on each pack How much do you have to pay to buy each pack?
\end{tcolorbox}

\begin{tcolorbox}[
    colback=gray!10, 
    colframe=gray!70,
    title={\Centering Chain-of-Draft}, 
    boxrule=0.5pt,
    arc=2pt,
    fonttitle=\bfseries,
    coltitle=white!50,
    listing only,
     breakable,
    listing options={
        basicstyle=\ttfamily\small,
        breaklines=true,
        escapeinside=@@,
    }
]
Answer: \\
1. Original cost: 76 dollars. \\
2. Discount: 25 dollars. \\
3. Subtract discount from cost. \\
\\
\#\#\#\# 51 dollars
\end{tcolorbox}

\begin{tcolorbox}[colback=blue!5!white, colframe=blue!60!black,breakable, title=\Centering Chain of Unconscious Thought]
Answer: \\
76 - 25 = 51 
\\
\#\#\#\# 51 dollars
\end{tcolorbox}

\noindent  Case 2 from SVAMP dataset is demonstrated as follows. 
\begin{tcolorbox}[
    colback=gray!10, 
    colframe=gray!70,
    title={\Centering Question}, 
    boxrule=0.5pt,
    arc=2pt,
    fonttitle=\bfseries,
    coltitle=white!50,
    listing only,
     breakable,
    listing options={
        basicstyle=\ttfamily\small,
        breaklines=true,
        escapeinside=@@,
    }
]
\ttfamily
{Q:} Tiffany was collecting cans for recycling. On monday she had 7 bags of cans. The next day she found 12 more bags worth of cans. How many more bags did she find on the next day than she had on monday?
\end{tcolorbox}

\begin{tcolorbox}[
    colback=gray!10, 
    colframe=gray!70,
    title={\Centering Concise Chain-of-Thought}, 
    boxrule=0.5pt,
    arc=2pt,
    fonttitle=\bfseries,
    coltitle=white!50,
    listing only,
     breakable,
    listing options={
        basicstyle=\ttfamily\small,
        breaklines=true,
        escapeinside=@@,
    }
]
Answer: \\
1. Determine the number of bags Tiffany had on Monday: 7 bags.\\
2. Determine the number of bags Tiffany found the next day: 12 bags.\\
3. Calculate the difference: 12 bags - 7 bags = 5 bags.\\
\#\#\#\# 5 bags
\end{tcolorbox}

\begin{tcolorbox}[
    colback=gray!10, 
    colframe=gray!70,
    title={\Centering Chain-of-Draft}, 
    boxrule=0.5pt,
    arc=2pt,
    fonttitle=\bfseries,
    coltitle=white!50,
    listing only,
     breakable,
    listing options={
        basicstyle=\ttfamily\small,
        breaklines=true,
        escapeinside=@@,
    }
]
Answer: \\
- Monday: 7 bags. \\
- Next day: 12 bags. \\
- Difference: 12 - 7. \\
- Calculate: 5 more bags. \\
\#\#\#\# 5
\end{tcolorbox}

\begin{tcolorbox}[colback=blue!5!white, colframe=blue!60!black,breakable, title=\Centering Chain of Unconscious Thought]
Answer: \\
12 - 7 = 5
\\
\#\#\#\# 5
\end{tcolorbox}

\noindent  Case 3 from MATHQA dataset is demonstrated as follows. 
\begin{tcolorbox}[
    colback=gray!10, 
    colframe=gray!70,
    title={\Centering Question}, 
    boxrule=0.5pt,
    arc=2pt,
    fonttitle=\bfseries,
    coltitle=white!50,
    listing only,
     breakable,
    listing options={
        basicstyle=\ttfamily\small,
        breaklines=true,
        escapeinside=@@,
    }
]
\ttfamily
{Q:} The population of a town is 8000 . it decreases annually at the rate of 20 \% What will be its population after 3 years ? A ) 5100 , B ) 4096 , C ) 5200 , D ) 5400 , E ) 5500
\end{tcolorbox}

\begin{tcolorbox}[
    colback=gray!10, 
    colframe=gray!70,
    title={\Centering Chain-of-Draft}, 
    boxrule=0.5pt,
    arc=2pt,
    fonttitle=\bfseries,
    coltitle=white!50,
    listing only,
     breakable,
    listing options={
        basicstyle=\ttfamily\small,
        breaklines=true,
        escapeinside=@@,
    }
]
Answer: \\
Initial population: 8000.  \\
Annual decrease: 20\%.  \\
Population after 1 year: 8000 * 0.8.  \\
Population after 2 years: 6400 * 0.8.  \\
Population after 3 years: 5120 * 0.8.  \\
Final population: 4096.  \\
\#\#\#\# B
\end{tcolorbox}

\begin{tcolorbox}[colback=blue!5!white, colframe=blue!60!black,breakable, title=\Centering Chain of Unconscious Thought]
Answer: \\
Population after 3 years = $8000 * (0.8)^3 = 4096$.  \\
\#\#\#\# B
\end{tcolorbox}

\end{document}